\documentclass[10pt,twocolumn,letterpaper]{article}

\usepackage{times}
\usepackage{epsfig}
\usepackage{graphicx}
\usepackage{amsmath}
\usepackage{amssymb}
\usepackage{pdfpages}
\usepackage{graphicx}
\usepackage{subcaption}
\usepackage{commath}
\usepackage{placeins}

\usepackage[pagebackref=true,breaklinks=true,letterpaper=true,colorlinks,bookmarks=false]{hyperref}

\begin{document}
\title{Improved Style Transfer by Respecting Inter-layer Correlations}

\author{Mao-Chuang Yeh and Shuai Tang\\
University of Illinois at Urbana Champaign\\
myeh2@illinois.edu, stang30@illinois.edu}


\twocolumn[{%
\renewcommand\twocolumn[1][]{#1}%
\maketitle
\begin{center}
    \centering
    \includegraphics[clip, trim=2.7cm 10.5cm 2.5cm 10.5cm, width=\textwidth]{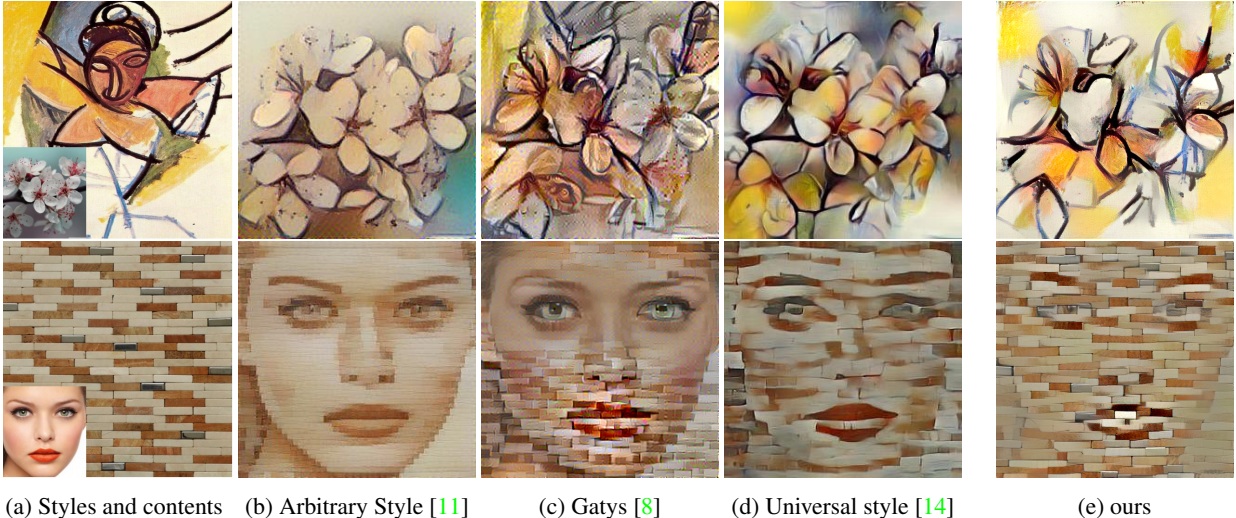}
    \captionof{figure}{The first column are style and content images; images in column 2,3,4 are results from \cite{huang2017arbitrary},  \cite{gatys2016image}, and \cite{UST}, which are reported in Y. Li, et al. \cite{UST}. Our results are at the last column, which use cross-layer gram matrices as style losses and are optimized on multiplicative loss between content and style.}
\label{fig:tiser}
\end{center}%
}]

\begin{figure*}[!h]
\centering
\includegraphics[width=0.98\linewidth]{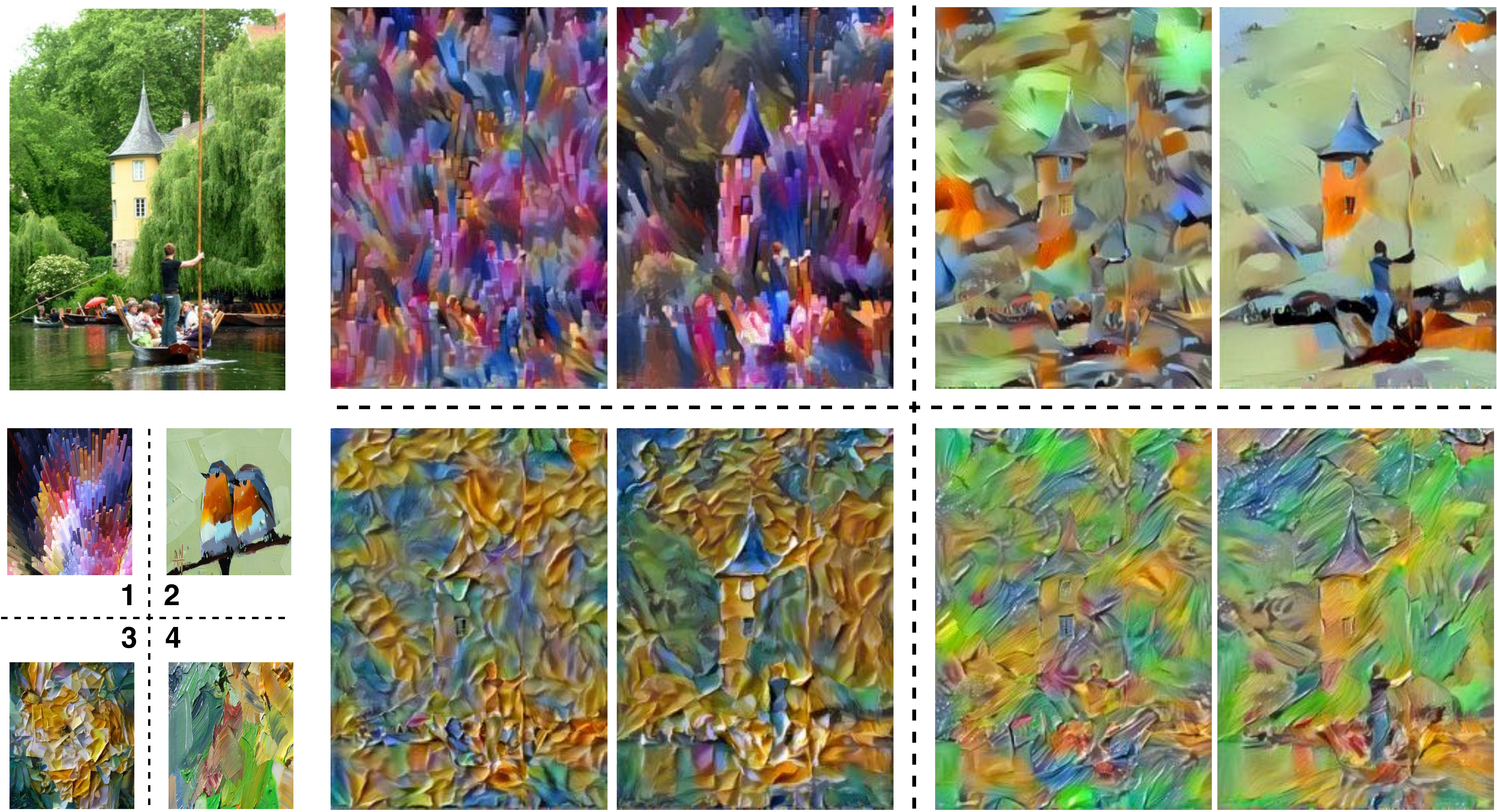}
\caption{\em Style transfer examples demonstrate the advantages of our method (\textbf{right} image in every example) compared to Gatys' method \cite{gatys2016image} (\textbf{left} image). 
We use cross-layer gram matrices (section \ref{sec:Cross}) and multiplicative loss (section \ref{sec:mul1}). Notice our method preserves prominent content boundaries (e.g. the tower and the boatman), while transferring patterns from the style image more completely. Our method excels at transferring style with long scale coherence (example 1,2); and preserves the appearance of material relief (example 3,4).
}
\label{fig:mulloss_comp}
\end{figure*}

\begin{abstract}
A popular series of style transfer methods apply a style to a content image by controlling mean and covariance of values in early layers of a feature stack. This is insufficient for transferring styles that have strong structure across spatial scales like, e.g., textures where dots lie on long curves. This paper demonstrates that controlling inter-layer correlations yields visible improvements in style transfer methods.  We achieve this control by computing cross-layer, rather than within-layer, gram matrices. We find that (a) cross-layer gram matrices are sufficient to control within-layer statistics. Inter-layer correlations improves style transfer and texture synthesis. The paper shows numerous examples on "hard" real style transfer problems (e.g. long scale and hierarchical patterns); (b) a fast approximate style transfer method can control cross-layer gram matrices; (c) we demonstrate that multiplicative, rather than additive style and content loss, results in very good style transfer. Multiplicative loss produces a visible emphasis on boundaries, and means that one hyper-parameter can be eliminated.   
\end{abstract}

\section{Introduction}

Style transfer methods apply the ``style'' from one example image to the ``content'' of another; for instance, one might render a camera image (the content) as a watercolor painting (the style). Recent work has shown that highly effective style transfer can be achieved by searching for an image such that early layers of CNN representation match  
the early layers of the style image and later layers match the later layers of a content image~\cite{gatys2016image}. Content matching is by comparing unit outputs at each location of feature map. But style matching is achieved by comparing summary statistics -- in particular, the gram matrix -- of the layers individually. Comparing gram matrices of individual layers ensures that small, medium and large patterns that are common in the style image appear with about the same frequency in the synthesized image, and that spatial co-occurences between these patterns are about the same in synthesized and style image. 

\begin{figure}[!t]
  \centering
  \begin{subfigure}[b]{0.3\linewidth}
    \includegraphics[width=\linewidth, height=\linewidth]{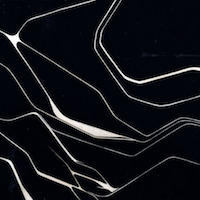}
  \end{subfigure}
  \begin{subfigure}[b]{0.3\linewidth}
    \includegraphics[width=\linewidth]{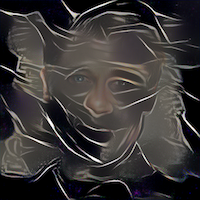}
  \end{subfigure}
  \begin{subfigure}[b]{0.3\linewidth}
    \includegraphics[width=\linewidth]{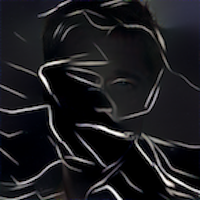}
  \end{subfigure}

  \begin{subfigure}[b]{0.3\linewidth}
    \includegraphics[width=\linewidth, height=\linewidth]{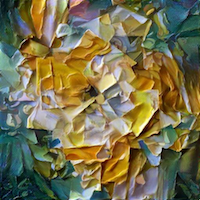}
  \end{subfigure}
  \begin{subfigure}[b]{0.3\linewidth}
    \includegraphics[width=\linewidth]{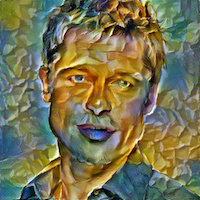}
  \end{subfigure}
  \begin{subfigure}[b]{0.3\linewidth}
    \includegraphics[width=\linewidth]{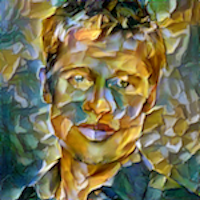}
  \end{subfigure}

  \begin{subfigure}[b]{0.3\linewidth}
    \includegraphics[width=\linewidth, height=\linewidth]{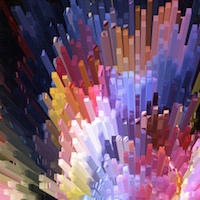}
  \end{subfigure}
  \begin{subfigure}[b]{0.3\linewidth}
    \includegraphics[width=\linewidth]{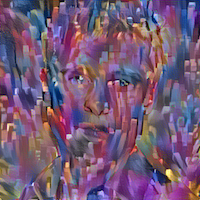}
  \end{subfigure}
  \begin{subfigure}[b]{0.3\linewidth}
    \includegraphics[width=\linewidth]{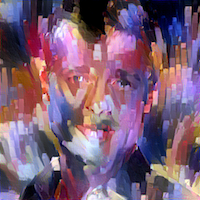}
  \end{subfigure}
  \begin{subfigure}[b]{0.3\linewidth}
    \includegraphics[width=\linewidth,
    height=\linewidth]{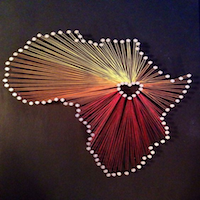}
\caption{Style}
  \end{subfigure}
  \begin{subfigure}[b]{0.3\linewidth}
    \includegraphics[width=\linewidth]{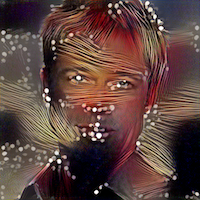}
\caption{Within-Layer}
  \end{subfigure}
  \begin{subfigure}[b]{0.3\linewidth}
    \includegraphics[width=\linewidth]{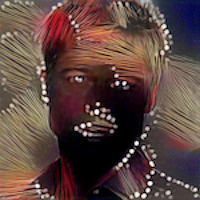}
\caption{Cross-Layer}
  \end{subfigure}  
  \caption{\em {\bf Left:} styles to transfer; {\bf center:} results using within-layer
    loss; {\bf right} results using cross-layer loss.  There are
    visible advantages to using the cross-layer loss. Note how cross-layer preserves large black areas (top row); 
creates an improved appearance of relief for the acrylic strokes (second row); preserves the overall structure of the
rods (third row); and ensures each string has a dot on each end (fourth row).
  }\label{fig:cf1}
\end{figure}

Novak and Nikulin noticed that across-layer gram matrices reliably produce improvement on style transfer. (\cite{novak2016improving}). However, their work was an exploration of variants of style transfer rather than a thorough study to gain insights on style summary statistics. There are reasons cross-layer terms produce improvements. In some styles, very long scale patterns are formed out of small components.  For instance, in Figure~\ref{fig:cf1}, small white spots are organized into long curves. Within-layer gram matrices are not well adapted to represent this phenomenon, as Figure~\ref{fig:cf1} shows. Generally, such hard styles occur where effects at short spatial scales are organized into longer scale structures. Such hard styles are strongly associated with physical materials (for instance, relief painting in Figure~\ref{fig:mulloss_comp}).  In this paper, we show that comparing cross-layer gram matrices -- which encode co-occurrences between (say) small and medium scale patterns --- produces improvements in style transfer for such styles. Furthermore, controlling cross-layer gram matrices also effectively controls pattern frequencies. 

\textbf{Our contributions:}
\begin{itemize}
\item We show that controlling cross-layer, rather than within-layer, gram matrices produces visible improvements in style transfer for many styles even though the cross-layer has less constraints than within-layer. This observation differs from the main claim of Novak and Nikulin, which suggests more layers(16layers) are needed for cross-layer gram matrix to improve within layer terms~\cite{novak2016improving}. Furthermore, they found reliable small improvements from cross-layer gram-matrices; in contrast, we argue that the method produces large, principled improvements, particularly for styles where inter-scale relations are important (Figure \ref{fig:tiser}, \ref{fig:mulloss_comp}).

\begin{figure}[!t]
  \centering
  \begin{subfigure}[b]{0.3\linewidth}
    \includegraphics[width=\linewidth, height=\linewidth]{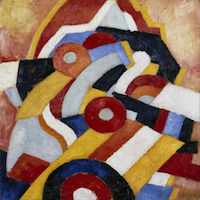}
  \end{subfigure}
  \begin{subfigure}[b]{0.3\linewidth}
    \includegraphics[width=\linewidth]{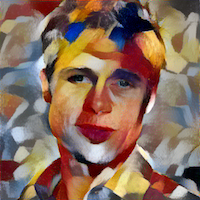}
  \end{subfigure}
  \begin{subfigure}[b]{0.3\linewidth}
    \includegraphics[width=\linewidth]{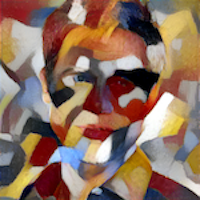}
  \end{subfigure}

  \begin{subfigure}[b]{0.3\linewidth}
    \includegraphics[width=\linewidth, height=\linewidth]{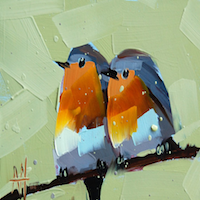}
  \end{subfigure}
  \begin{subfigure}[b]{0.3\linewidth}
    \includegraphics[width=\linewidth]{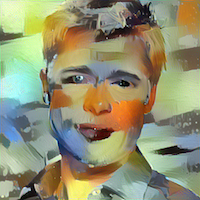}
  \end{subfigure}
  \begin{subfigure}[b]{0.3\linewidth}
    \includegraphics[width=\linewidth]{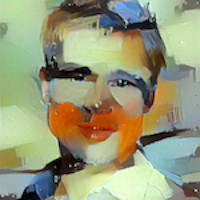}
  \end{subfigure}

  \begin{subfigure}[b]{0.3\linewidth}
    \includegraphics[width=\linewidth, height=\linewidth]{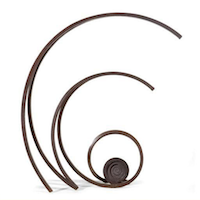}
  \end{subfigure}
  \begin{subfigure}[b]{0.3\linewidth}
    \includegraphics[width=\linewidth]{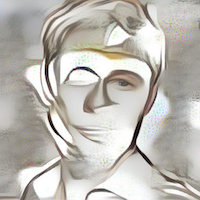}
  \end{subfigure}
  \begin{subfigure}[b]{0.3\linewidth}
    \includegraphics[width=\linewidth]{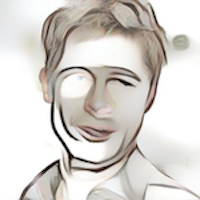}
  \end{subfigure}  
  
  \begin{subfigure}[b]{0.3\linewidth}
    \includegraphics[width=\linewidth, height=\linewidth]{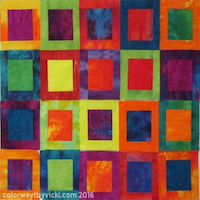}
\caption{Style}
  \end{subfigure}
  \begin{subfigure}[b]{0.3\linewidth}
    \includegraphics[width=\linewidth]{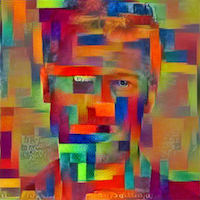}
\caption{Within-Layer}
  \end{subfigure}
  \begin{subfigure}[b]{0.3\linewidth}
    \includegraphics[width=\linewidth]{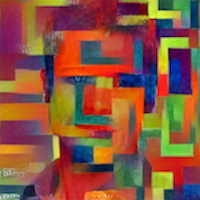}
\caption{Cross-Layer}
  \end{subfigure} 
    \caption{\em {\bf Left:} styles to transfer; {\bf center:} results using within-layer
    loss; {\bf right:} results using cross-layer loss.  There are
    visible advantages to using the cross-layer loss. Note how cross-layer preserves the shape of the abstract color blocks (top row); 
avoids smearing large paint strokes (second row); preserves the overall structure of the curves as much as possible
(third row); and produces color blocks with thin boundaries (fourth row).
  }\label{fig:cf2}
\end{figure}

\item We show that universal style transfer (UST) method can adapt to cross-layer gram matrices, consequently improving style transfer.
\item We demonstrate that loss multiplication method often produce better looking style transfers. We claim that multiplicative loss has stronger capability to encourage the stylized image to preserve prominent boundaries in content image geometry than additive loss does. 
\end{itemize}



\section{Related work}
Bilinear models are capable of simple image style transfer~\cite{Tenenbaum2000} by factorizing style and content representations, but non-parametric methods like patch-based texture synthesis can deal with much more complex texture fields~\cite{Efros2001}.  Image analogies
use a rendering of one image in two styles to infer a mapping from a content image to a stylized image~\cite{Hertzmann2001}. Researchers have been looking for versatile parametric methods to control style patterns at different scales to be transferred. Adjusting filter statistics is known to yield texture synthesis~\cite{debonet,simoncelli}.  Gatys et al. demonstrated that producing neural network layers with particular summary statistics (i.e Gram matrices) yielded effective texture synthesis~\cite{NIPS2015_5633}. In a following paper, Gatys et al. achieved style transfer by searching for an image that satisfies both style texture summary statistics and content constraints~\cite{gatys2016image}. This work has been much elaborated. The search can be replaced with a regression (at one scale~\cite{Johnson2016Perceptual}; at multiple scales~\cite{wang2016multimodal}; with cached~\cite{chen2017stylebank} or learned~\cite{dumoulin2016learned} style representations) or a decoding process that allows efficient adjusting of statistics~\cite{UST}. Search can be sped up with local matching methods~\cite{chen2016fast}. Methods that produce local maps (rather than pixels) result in photorealistic style transfer~\cite{Shih2014,Luan2017}. Style transfer can be localized to masked regions~\cite{gatys2016controlling}. The criterion of matching summary statistics is a Maximum Mean Discrepancy condition~\cite{li2017demystifying}. Style transfer can be used to enhance sketches~\cite{champandard2016semantic}. 

Novak and Nikulin search a range of variant style transfer methods, including cross-layer gram matrices. However, their primary suggestions are adding more layers for more features, and shifting activations such that the number of zero entries in gram matrix is reduced. They don't pursue on cross-layer gram matrices nor explain its results. They experiment on a long chain of cross-layer gram matrices but do not identify what the improvements are or extend the method to fast style transfer~\cite{novak2016improving}.  There is a comprehensive review in~\cite{jing2017neural}.

\section{Within layer gram matrix for style transfer}

Gatys et al. \cite{gatys2016image} finds an image where early layers of a CNN representation match the lower layers of the style image and higher layers match the higher layers of a content image.  We review the original work of Gatys et al. in detail.  Write $I_{s}$ (resp. $I_{c}$, $I_{n}$)  for the style (resp. content, new) image,
and $\alpha$ for some parameter balancing style and content losses ($L_s$ and $L_c$ respectively).
We obtain $I_{n}$ by optimizing
\[
L_c(I_{n}, I_{c})+\alpha L_s(I_{n}, I_{c})
\]
Losses are computed on a network representation, with $L$ convolutional layers, where the $l$'th layer
produces a feature map $f^l$ of size $H^l \times W^l \times C^l$ (resp. height, width, and channel number). We partition
the layers into three groups (style, content and irrelevant). Then we reindex the spatial variables (height and width) and
write $f^l_{k,p}$ for the response of the $k$'th channel at the  $p$'th location in the $l$'th convolutional layer. The
content loss $L_c$ is 
\[
L_c(I_{n}, I_{c}) = \frac{1}{2}\sum_{c} \sum_{k,p} \norm{f^c_{k,p}(I_{n}) - f^c_{k,p}(I_{c})}^2
\]
(where $c$ ranges over content layers). The style loss is depends on within-layer gram matrices.  Write
\[
G_{ij}^l(I) = \sum_p \left[f_{i,p}^l(I)\right]\left[f_{j,p}^l(I)\right]^{T}
\]
and $w_l$ for the weight applied to the $l$'th layer.  Then 
\[
L_s^l(I_{n}, I_{s}) = \frac{1}{4{P^l}^2{K^l}^2}\sum_{s}w_l \sum_{i,j}\norm{G^s_{ij}(I_{n})-G^s_{ij}(I_{s})}^2
\]
where $s$ ranges over style layers. Gatys et al. use Relu1\_1, Relu2\_1, Relu3\_1, Relu4\_1, and Relu5\_1 as style layers, and layer Relu4\_2 
for the content loss, and search for $I_{n}$ using L-BFGS~\cite{liu1989limited}.  {\bf Notation:} From now on, we write R51 for Relu5\_1, etc.

\section{The cross layer gram matrix}\label{sec:Cross}

Now consider layer $l$ and $m$, both style layers, with decreasing spatial resolution. 
Write $\uparrow f^{m}$ for an upsampling of  $f^m$ to $H^l\times W^l \times K^l$, and consider
\[
G_{ij}^{l,m}(I) = \sum \left[ f_{i,p}^l(I)\right]\left[\uparrow {f}_{j,p}^{m}(I)\right]^{T}.
\]
as the cross-layer gram matrix, We can form a style loss
\[
L_s(I, I_{s}) = \sum_{(l, m)\in {\cal L}} w^{l}\sum_{ij} \norm{G^{l,m}_{ij}(I)-G^{l,m}_{ij}(I_s)}^2
\]
(where ${\cal L}$ is a set of pairs of style layers).   We can substitute this loss into the original style loss, and
minimize as before.   This construction has a variety of interesting properties which we will investigate later.

{\bf Style layer pairs:}  In principle, any set of pairs can be used.  We have investigated a {\em pairwise descending}
strategy, where one constrains each layer and its successor (i.e. (R51, R41); (R41, R31); etc) and an {\em all distinct pairs}
strategy, where one constrains all pairs of distinct layers.

{\bf Pattern management across scales:}  Controlling within-layer gram matrices by proper weighting ensures that the statistics of patterns
at a particular scale are ``appropriate''. However, we speculate -- and our experimental results seem to confirm --
that one can get these statistics right without having desirable weighting relations across scales.  Inter-layer gram matrices
require that phenomena at one scale are correlated to those at the next scale appropriately. In other words, carefully controlling weights for each layer's style loss is not necessary in cross-layer gram matrix scenario. 

{\bf Number of constraints:} Cross-layer gram matrices control considerably fewer parameters than within layer gram
matrices. For a pairwise descending strategy, we have four cross-layer gram matrices, leading to 
control of $64\times
128+128\times 256+256\times 512+512\times 512 = 434176 $ parameters; compare within layer gram matrices, which control 
$64^2+128^2+256^2+2\times512^2 = 610304$ parameters.  It may seem that there is less constraint on style.  Experiment
suggests our method produces visible improved results, meaning that many of the parameters controlled by within-layer
gram matrices have no particular effect on the outcome. 



\begin{figure}[!t]
\centering
\includegraphics[width=\linewidth]{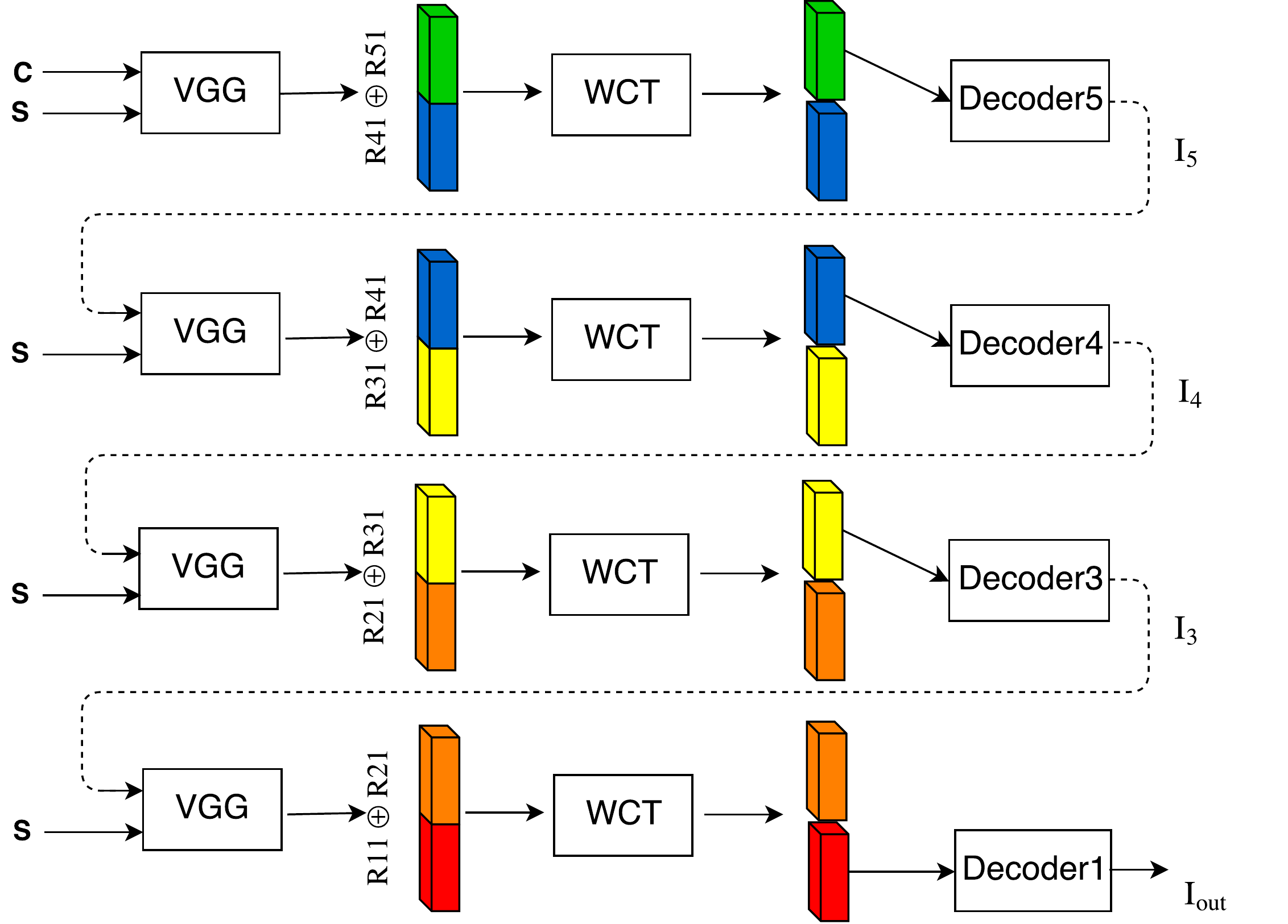}
\caption{\em Fast universal cross-layer transfer (FCT). We use similar procedure as in Li et al. \cite{UST}, a pair of convlutional features (e.g. R11 and R21) are reshaped and concatenated before performing transformation. The transformed feature is then split up and only one layer is fed into the decoder. We use off-shelf decoders from \cite{UST}.  }
\label{fig:FCT_flow}
\end{figure}

\subsection{Fast Universal Cross-layer Transfers}

Li et al. use signal whitening and coloring to implement a fast version of style transfer using a VGG encoder~\cite{UST}.  Their procedure takes
the R51 layer from the content image, then applies an affine transformation (by whitening, coloring, and matching means) to match the gram matrix of the corresponding layer computed for the style image.  The resulting layer is decoded to an image through one of five pre-trained image reconstruction decoder networks. The R41 layer produced by this image is again affine transformed to match the gram matrix of the corresponding layer computed for the style image.  This layer is then again decoded to an image.  The process continues until the affine transformed R11 layer is decoded to an image, which is retained.

This procedure is easily extended to cross-layer gram matrices (Figure~\ref{fig:FCT_flow}). We start by choosing sequence of sets of layer covariances to control.  The simple scheme is individual, controlling (R51), (R41), (R31), (R21), (R11). An alternative scheme is
pairwise descending, where one controls (R51, R41); (R41, R31); (R31, R21); and (R21, R11).  Another scheme is descending, where one controls (R51, R41, R31, R21, R11); (R41, R31, R21, R11); (R31, R21, R11); etc. we then start the first set of relevant layers from the content image (i.e. (R51, R41) for pairwise; (R51, ... R11) for
descending).  Construct a gram matrix from this {\em set} of layers, upsampling as required. 
Apply an affine transformation to match the gram matrix of the corresponding {\em set} of layers for the
style image, then decode the resulting layers to an image.  Pass this image through VGG, recover the next set of control
layers from the result, apply an affine transformation to match the gram matrix of the corresponding {\em set} of layers for the
style image, then decode the resulting layers to an image.  Proceed until the R11 layer is decoded to an image, and use
that image.   Note that this approach controls both within layer and between layer statistics, as the relevant gram
matrices have within layer gram matrices as diagonal blocks, and between layer gram matrices as off-diagonal blocks.
\begin{figure}[!t]
\centering
\small 
\begin{subfigure}[t]{0.3\linewidth}
    \includegraphics[width=\linewidth]{AddCrossGatys_content122_style_style48.png}
    \includegraphics[width=\linewidth]{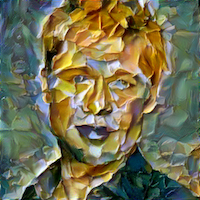}
\end{subfigure}
\begin{subfigure}[t]{0.3\linewidth}
    \includegraphics[width=\linewidth]{AddCrossGatys_content122_style_style6.png}
    \includegraphics[width=\linewidth]{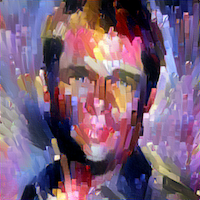}
\end{subfigure}
\begin{subfigure}[t]{0.3\linewidth}
    \includegraphics[width=\linewidth]{AddCrossGatys_content122_style_style117.png} 
    \includegraphics[width=\linewidth]{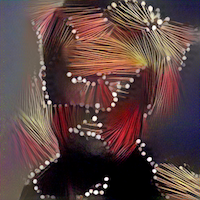}
\end{subfigure}
\caption{\em 
 Multiplicative loss produces good style transfer results.
{\bf Top row:} style transfers using cross layer gram matrices and additive loss, with a good choice of $\alpha$. {\bf
  Bottom row:} style transfers using cross layer gram matrices and multiplicative loss, where no choice of $\alpha$ is required. Notice the emphasis of  content outline in the multiplicative loss images. 
}
  \label{fig:MulCG}
\end{figure}

\subsection{Loss multiplication}\label{sec:mul1}

Style transfer methods require a choice of parameter, $\alpha$, to balance the style and content losses.   The value is
typically chosen by eye, which is unsatisfying.  A natural alternative to adding the losses is to multiply them; in this
case, no parameter is needed, and we can form
\[
L^m(I_n) = L_c(I_n, I_c) *  L_s(I_n, I_s).
\]
Multiplicative loss tends to emphasize strong boundaries in the content image( Figure~\ref{fig:MulCG}). We believe this is because style loss is always large, so that minimization will force down large differences (which are large difference in values between stylized image and content image) in the content layer.  Our experimental results suggest that this approach is successful (section~\ref{sec:mul}).  The effect is quite prominent (Figure~\ref{fig:MulCG}, Figure~\ref{fig:scale2}, Figure~\ref{fig:scale3} ), multiplicative loss has significant advantage of reducing the number of parameters that need to be searched over to produce useful results.  Figure~\ref{fig:mulloss_comp} shows style transfer results using cross-layer gram matrices and multiplicative loss, we observe distinguishable improvement over Gatys' method in preserving content boundaries.

\begin{figure*}[!t]
\centering
\small 
\begin{subfigure}[t]{0.15\textwidth}
	
	\includegraphics[width=\linewidth]{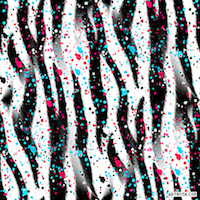}
	\includegraphics[width=\linewidth]{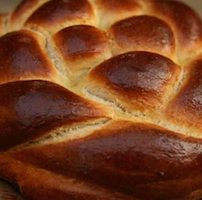}
	\includegraphics[width=\linewidth]{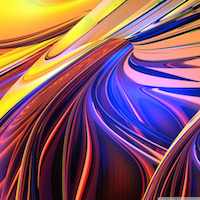}
	\includegraphics[width=\linewidth]{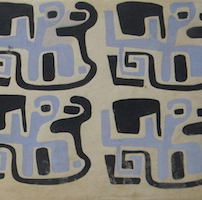}
    \includegraphics[width=\linewidth]{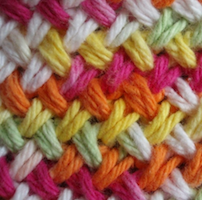}
    \caption{Styles}
\end{subfigure}
\begin{subfigure}[t]{0.15\textwidth}
	
	\includegraphics[width=\linewidth]{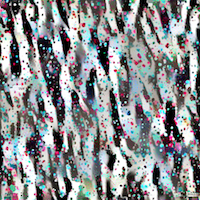}
	\includegraphics[width=\linewidth]{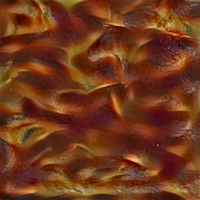}
	\includegraphics[width=\linewidth]{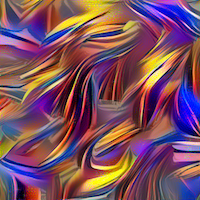}
	\includegraphics[width=\linewidth]{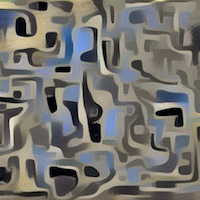}  
    \includegraphics[width=\linewidth]{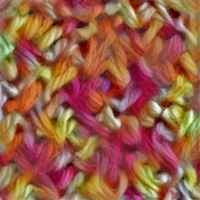}
    \caption{Within-layers}
\end{subfigure}
\begin{subfigure}[t]{0.15\textwidth}
	
	\includegraphics[width=\linewidth]{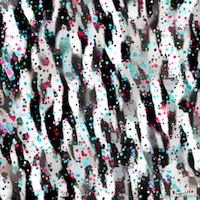}
	\includegraphics[width=\linewidth]{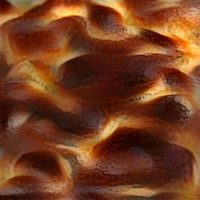}
	\includegraphics[width=\linewidth]{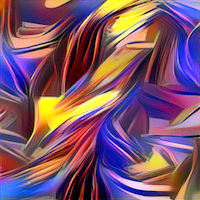}
	\includegraphics[width=\linewidth]{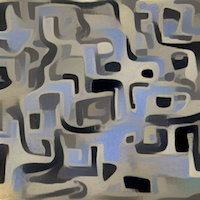}
    \includegraphics[width=\linewidth]{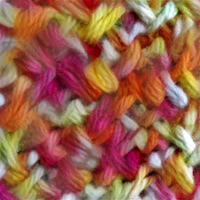}
    \caption{CG}
\end{subfigure}
\begin{subfigure}[t]{0.15\textwidth}

	\includegraphics[width=\linewidth]{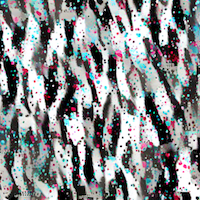}
    \includegraphics[width=\linewidth]{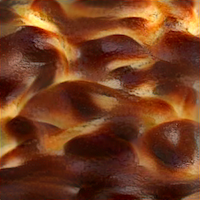}
    \includegraphics[width=\linewidth]{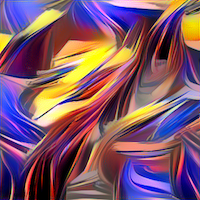}
    \includegraphics[width=\linewidth]{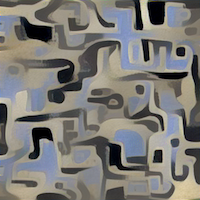}
    \includegraphics[width=\linewidth]{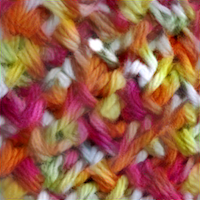}
    \caption{more CGs}
\end{subfigure}
\begin{subfigure}[t]{0.153\textwidth}
	
	\includegraphics[width=\linewidth]{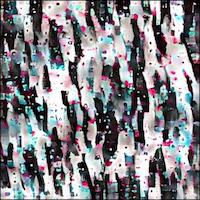}
	\includegraphics[width=\linewidth]{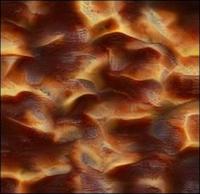}
	\includegraphics[width=\linewidth]{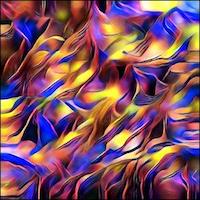}
	\includegraphics[width=\linewidth]{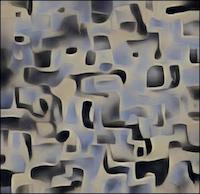}
    \includegraphics[width=\linewidth]{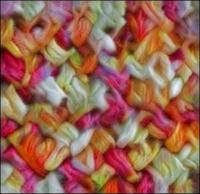}
    \caption{WCT}
\end{subfigure}
\begin{subfigure}[t]{0.153\textwidth}
	
	\includegraphics[width=\linewidth]{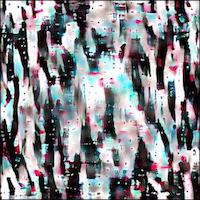}
	\includegraphics[width=\linewidth]{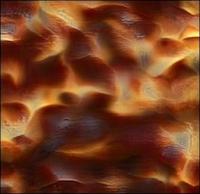}
	\includegraphics[width=\linewidth]{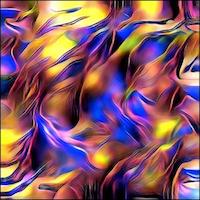}
	\includegraphics[width=\linewidth]{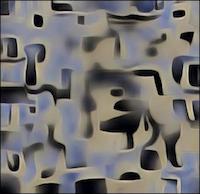}
    \includegraphics[width=\linewidth]{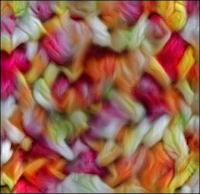}
    \caption{FCT}
\end{subfigure}
\caption{Texture synthesis comparison: Except the first column as
  style, the rest of columns from left to right are respectively generated by
  within-layer gram matrix, CG (cross-layer gram matrices), more CG (all cross-layer gram matrices between R51,R4,R31,R21,R11 are considered), WCT, and FCT. We can see that
  either in Gatys vs ours or WCT vs FCT, the cross-layer gram
  matrix indeed shows the improvement on texture patterns. }
\label{fig:texture}
\end{figure*}

We find one trick to improve transfer results using multiplicative loss by shifting the mean when creating the new image to optimized, we recommend this shift should be the channel mean of style image.
\section{Results}

\subsection{Experimental details}

We use VGG-19 for both style transfer and texture synthesis.  We use R11, R21, R31, R41, and R51 for style(texture) loss, and R42 for the content loss for style transfer. In loss optimization, if it not specified, all stylized images start from Gaussian noise image and optimized with LBFGS.

\begin{figure}[!tbhp]
  \centering
  \begin{subfigure}[b]{\linewidth}
    \includegraphics[width=\linewidth]{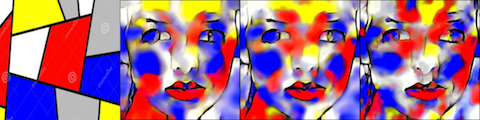}
     \caption{\em Our method shows better color grouping in the stylized image.}
  \end{subfigure}
  \begin{subfigure}[b]{\linewidth}
    \includegraphics[width=\linewidth]{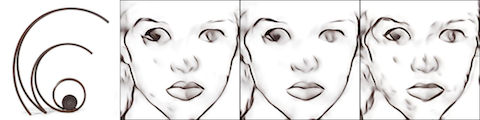}
    \caption{\em  Many black spots in original WCT, which is not observed in our method.}
  \end{subfigure}
  \begin{subfigure}[b]{\linewidth}
    \includegraphics[width=\linewidth]{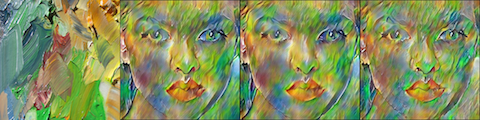}
    \caption{\em  Ours improved the color contrast, because cross-layer gram matrices preserve longer scale color pattern.}
  \end{subfigure}
  \begin{subfigure}[b]{\linewidth}
    \includegraphics[width=\linewidth]{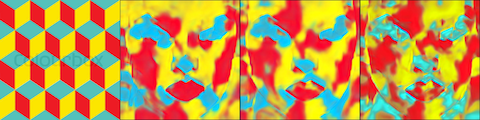}
    \caption{\em Note our method does not have the blue color shift present in WCT.}
  \end{subfigure}
  \begin{subfigure}[b]{\linewidth}
    \includegraphics[width=\linewidth]{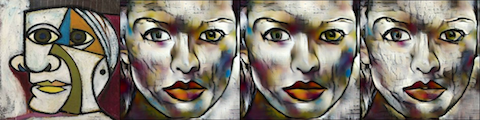}
    \caption{\em WCT has many artificial pattern which is not seen in original style image, and ours largely reduce it.}
  \end{subfigure}
  \begin{subfigure}[b]{\linewidth}
    \includegraphics[width=\linewidth]{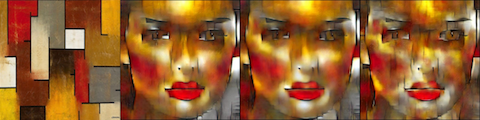}
    \caption{\em Color blocks are better organized in ours.}
  \end{subfigure}
  \caption{\em In each row, {\bf first:} the style image; {\bf second:} transfer using FCT with descending sequences
    (i.e. (R51, R41, R31, R21, R11); (R41, R31, R21, R11); (R31, R21, R11); etc); {\bf third:} transfer using FCT with pairwise descending sequences (i.e. (R51, R41); (R41, R31); (R31, R21); and (R21, R11)); {\bf fourth} transfer using
    WCT \protect \cite{UST}
  \label{fig:WCT1}}
\end{figure}

\subsection{Texture synthesis} 
Cross-layer gram matrix control applies to texture synthesis since style loss \cite{gatys2016image} was first introduced as "texture" loss in \cite{NIPS2015_5633}.  We now omit the content loss, and seeks a minimum of style loss alone.  We show texture synthesis results, which highlights the method's ability to manage long spatial correlations.  We controlled R51, R41, R31, R21, R11 for comparison with our style transfer results.  Our synthesis starts from an image which has the mean color of the texture image. 
As Figure \ref{fig:texture} shows, synthesized textures have better long scale coherence.    For the universal texture synthesis, we followed Li et al.  as starting from zero-mean Gaussian noise, run the multi level pipeline 3 times for better results.


\subsection{Style transfer}\label{sec:mul}
{\bf Cross-layer vs. within-layer style loss:}  Figures~\ref{fig:cf1},~\ref{fig:cf2} compare style transfers using within-layer gram matrices
and cross-layer gram matrices with a pairwise descending strategy. Cross-layer gram matrices are particularly good at
preserving relations between effects, as the detail in figure~\ref{fig:cf2} shows.

{\bf Multiplicative loss:} The multiplicative loss often produces visual pleasing style transfer results by showing better style pattern arrangement at same time keeping the outline of content relatively intact, so that the generated image preserves the perceptual meaning of the content while showing coherent style patterns(Figure~\ref{fig:MulCG}); More examples are present in supplementary materials.  


\begin{figure}[!b]
\centering
\small 
\begin{subfigure}[t]{0.3\linewidth}
    \includegraphics[width=\linewidth]{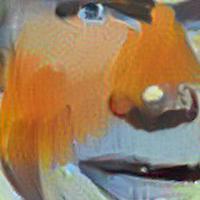}
    \includegraphics[width=\linewidth]{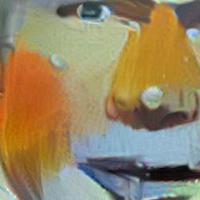}
\end{subfigure}
\begin{subfigure}[t]{0.3\linewidth}
    \includegraphics[width=\linewidth]{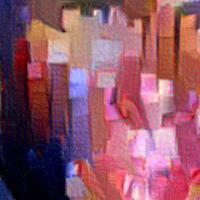}
    \includegraphics[width=\linewidth]{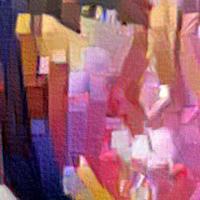}
\end{subfigure}
\begin{subfigure}[t]{0.3\linewidth}
    \includegraphics[width=\linewidth]{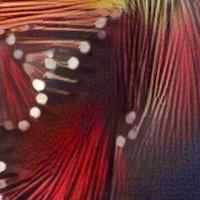}
    \includegraphics[width=\linewidth]{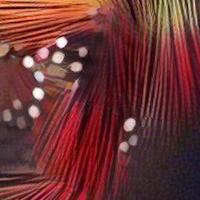}
\end{subfigure}
\caption{\em All pairs distinct cross-layer style transfer yields somewhat better results than descending pairs.  {\bf
    Top row:} cross-layer style transfer using descending pairs (i.e. (R51, R41); (R41, R31); (R31, R21); (R21, R11)).
  {\bf Bottom row:}  cross-layer style transfer using all pairs distinct (i.e  all distinct pairs from R51...R11).
There are fewer bubbles; color localization and value is improved; and line breaks are fewer.
  \label{fig:CGALL}}
\end{figure}

{\bf Pairwise descending vs all pairs distinct:}  All pairs distinct cross-layer style transfer seems to produce
improvements over pairwise descending (Figure~\ref{fig:CGALL}).  This is in some contrast to Novak and Nikulin's findings
(\cite{novak2016improving}, p5), which suggest ``tying distant ... layers produces poor results''.

{\bf FCT vs WCT:}  Fast universal cross-layer transfer (FCT) works visually better than the original WCT method of Li et al. ~\cite{UST}, as Figure~\ref{fig:WCT1} shows.  However, FCT has some of the same difficulties that WCT has.  Both methods have
difficulty reproducing crisp subshapes in styles. 

\begin{figure}[!h]
\centering

\includegraphics[clip, trim=3.5cm 9.4cm 8cm 4.5cm, width=\linewidth]{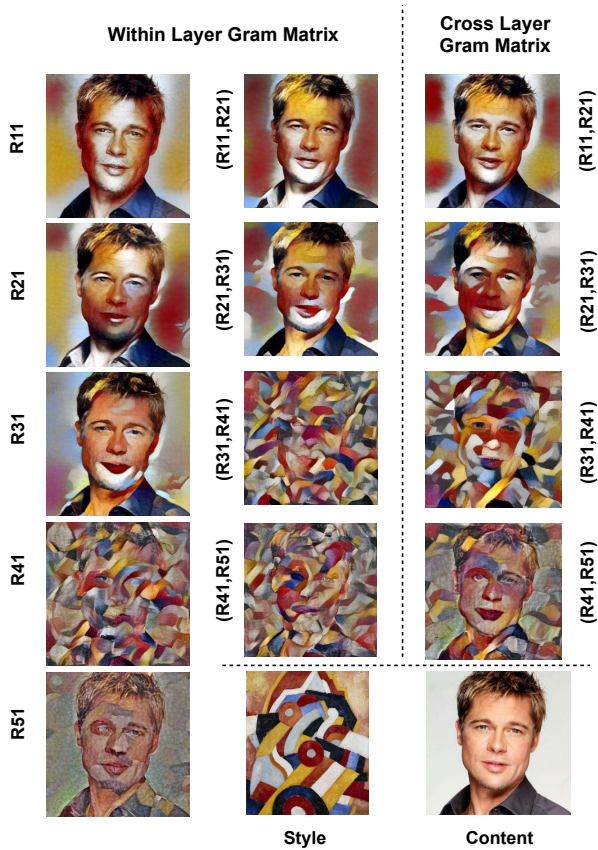}

\caption{\em This figure shows what happens when one controls only one (or one pair) of layers with the style loss.
{\bf Left:} controlling a single layer, with a within-layer gram matrix.  {\bf Center:} controlling two
layers in sequence, but each with a within-layer gram matrix.  {\bf Right:} controlling a two layers
in sequence, but using only a cross-layer gram matrix.  Notice that, as one would expect, controlling
cross-layer gram matrices results in more pronounced effects and a wider range of spatial scales of effect.
Furthermore, in comparison to controlling  a pair of within-layer gram matrices, one is controlling fewer
parameters.}
\label{fig:layer_wise}
\end{figure}

{\bf Individual style loss control:} When one controls style loss using a single layer (or a single pair of layers). We can clearly see how they effect stylized images (Figure~\ref{fig:layer_wise}). We observe higher level style loss shows stronger control over long scale patterns from style images, this is in agreement with similar observations in \cite{gatys2016image}. We also found that cross-layer gram matrices have stronger ability in preserving prominent boundaries of content images while display equal or better control over long scale style patterns compared to the same level within-layer gram matrices.  

\begin{figure}[!h]
  \centering 
	\begin{subfigure}[b]{0.157\linewidth}
    \includegraphics[width=\linewidth]{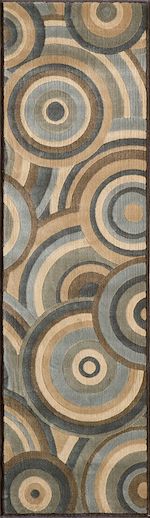}
    \caption{Style}
  \end{subfigure}
  \begin{subfigure}[b]{0.27\linewidth}
    \includegraphics[width=\linewidth, height=\linewidth]{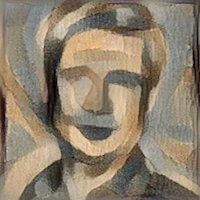}
    \includegraphics[width=\linewidth]{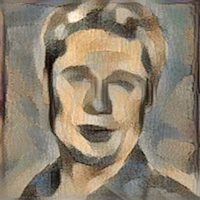}
    \caption{Style size 768}
  \end{subfigure}
  \begin{subfigure}[b]{0.27\linewidth}
    \includegraphics[width=\linewidth]{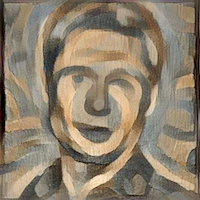}
    \includegraphics[width=\linewidth, height=\linewidth]{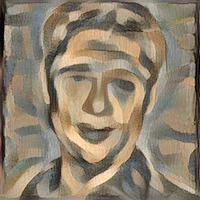}
    \caption{Style size 512}
  \end{subfigure} 
  \begin{subfigure}[b]{0.27\linewidth}
    \includegraphics[width=\linewidth]{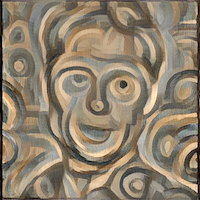}
    \includegraphics[width=\linewidth]{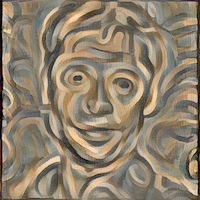}
    \caption{Style size 256}
  \end{subfigure}
  \caption{\em  Each row of stylized images shows a  transfer with the same style, but where the style image has been cropped to
    different sizes (style elements are {\em large} (=edge length 768), {\em medium} (=edge length 512) and {\em small}
    (=edge length 256), reading left to right).  The first row shows cross layer loss, the second row within layer loss.
    Note that, when style elements are large, the cross-layer loss is better at preserving their structure (e.g.,
  the large circles have fewer wiggles, etc.). Loss is multiplicative, notice the emphasis on outlines from multiplicative loss.
  }
  \label{fig:scale2}
\end{figure}

\begin{figure}[!ht]
  \centering

  \begin{subfigure}[b]{0.3\linewidth}
    \includegraphics[width=\linewidth]{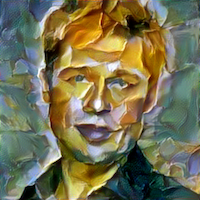}
  \end{subfigure}
  \begin{subfigure}[b]{0.3\linewidth}
    \includegraphics[width=\linewidth]{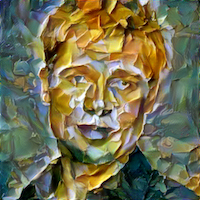}
  \end{subfigure}
  \begin{subfigure}[b]{0.3\linewidth}
    \includegraphics[width=\linewidth]{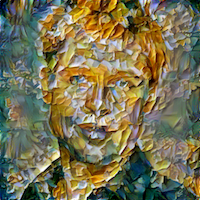}
  \end{subfigure}

  \begin{subfigure}[b]{0.3\linewidth}
    \includegraphics[width=\linewidth]{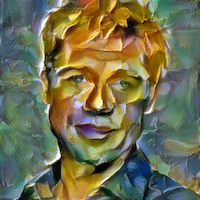}
    \caption{Style size 768}
  \end{subfigure}
  \begin{subfigure}[b]{0.3\linewidth}
    \includegraphics[width=\linewidth]{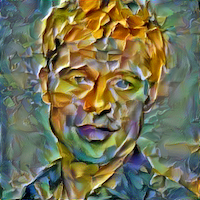}
    \caption{Style size 512}
  \end{subfigure}
  \begin{subfigure}[b]{0.3\linewidth}
    \includegraphics[width=\linewidth]{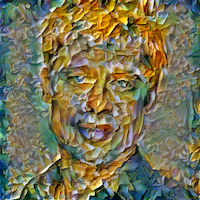}
    \caption{Style size 256}
  \end{subfigure}
\caption{\em  Each row shows a style transfer with the same style, but where the style image has been cropped to
    different sizes (style elements are {\em large} (=edge length 768), {\em medium} (=edge length 512) and {\em small}
    (=edge length 256), reading left to right).  The first row shows cross layer loss, the second row within layer loss.
    Note that, when style elements are large, the cross-layer loss is better at preserving their structure (e.g., the long scale color coherence is preserved, and
  the large paint strokes have more detail and more relief etc.)  Loss is multiplicative, notice the emphasis on outlines from multiplicative loss.
 }
  \label{fig:scale3}
\end{figure}

{\bf Scales:}  A crop of the style image will effectively result in transferring larger style elements.  We expect that,
when style elements are large compared to the content, cross-layer methods will have a strong advantage because they
will be better able to preserve structural relations that make up style elements.   Qualitative evidence supports this
view (Figure~\ref{fig:scale2} and Figure~\ref{fig:scale3}).

\section{Conclusion}

 \textbf{Cross-layer gram matrix} creates summary statistics that captures the correlation between different layers; higher layers can guide lower layers the most likely location for feature activations through the spacial product of forming cross-layer gram matrix. Therefore, we expect cross-layer gram matrices performs better especially on long scale patterns. Our experiments prove this point. The cross-layer gram matrix has less constraint but better style control than within-layer gram matrix. 

\textbf{Fast Universal Cross-layer Style Transfer} successfully unifies the Universal style transfer with our inter-layer statistics, and indeed shows some intrinsic difference in both style transfer and texture synthesis. 

\textbf{Multiplicative style loss} not only simplifies the style weight searching by eliminating one hyperparameter, but also emphasizes the boundary of content object even when strong boundaries information is present in style summary statistics. It provides better style quality in terms of preserving content shape and keeping long style coherence. More examples are present in the supplementary materials.

\section*{Acknowledgement}
 We give our deep thanks to Mao-Chuang's supervisor, Professor David Forsyth who provides many helpful suggestions in research and helps us writing the paper. We also thank Jason Rock for his help on setting up faster style transfer and other related recommendations. Besides, Anand Bhattad also gives us a lot of useful suggestions about style transfer. We really thank his kind help.

{\small
\bibliographystyle{ieee}
\bibliography{Arxive}
}

\end{document}